\documentclass[runningheads]{llncs}

\usepackage{eccv}

\usepackage{eccvabbrv}
\usepackage{algorithm}
\usepackage{algpseudocode}
\usepackage{graphicx}
\usepackage{booktabs}
\usepackage{multirow}        
\usepackage[table]{xcolor}   

\usepackage[accsupp]{axessibility}  

\usepackage{hyperref}

\usepackage{orcidlink}

\begin{document}

\title{T-DuMpRa: Teacher-guided Dual-path Multi-prototype Retrieval Augmented framework for fine-grained medical image classification} 

\titlerunning{T-DuMpRa}

\author{Zixuan Tang\inst{1} \and
Shen Zhao\inst{1}}

\authorrunning{Z.~Tang et al.}

\institute{School of Intelligent Systems Engineering, Sun Yat-sen University, Shenzhen, China\\
\email{\{tangzx23@mail2,zhaosh35@mail\}.sysu.edu.cn}}

\maketitle

\begin{abstract}
Fine-grained medical image classification is challenged by subtle inter-class variations and visually ambiguous cases, where confidence estimates often exhibit uncertainty rather than being overconfident. In such scenarios, purely discriminative classifiers may achieve high overall accuracy yet still fail to distinguish between highly similar categories, leading to miscalibrated predictions. We propose \textbf{T-DuMpRa}, a teacher-guided dual-path multi-prototype retrieval-augmented framework, where discriminative classification and multi-prototype retrieval jointly drive both training and prediction. During training, we jointly optimize cross-entropy and supervised contrastive objectives to learn a cosine-compatible embedding geometry for reliable prototype matching. We further employ an exponential moving average (EMA) teacher to obtain smoother representations and build a multi-prototype memory bank by clustering teacher embeddings in the teacher embedding space. Our framework is \textbf{plug-and-play}: it can be easily integrated into existing classification models by constructing a compact prototype bank, thereby improving performance on visually ambiguous cases. At inference, we combine the classifier’s predicted distribution with a similarity-based distribution computed via cosine matching to prototypes, and apply a conservative confidence-gated fusion that activates retrieval only when the classifier's prediction is uncertain and the retrieval evidence is decisive and conflicting, otherwise keeping confident predictions unchanged. On HAM10000 and ISIC2019, our method yields 0.68\%-0.21\% and 0.44\%-2.69\% improvements on 5 different backbone. And visualization analysis proves our model can enhance the model’s ability to handle visually ambiguous cases.
\end{abstract}

\section{Introduction}

\begin{figure}[tb]
    \centering
    \includegraphics[width=\textwidth]{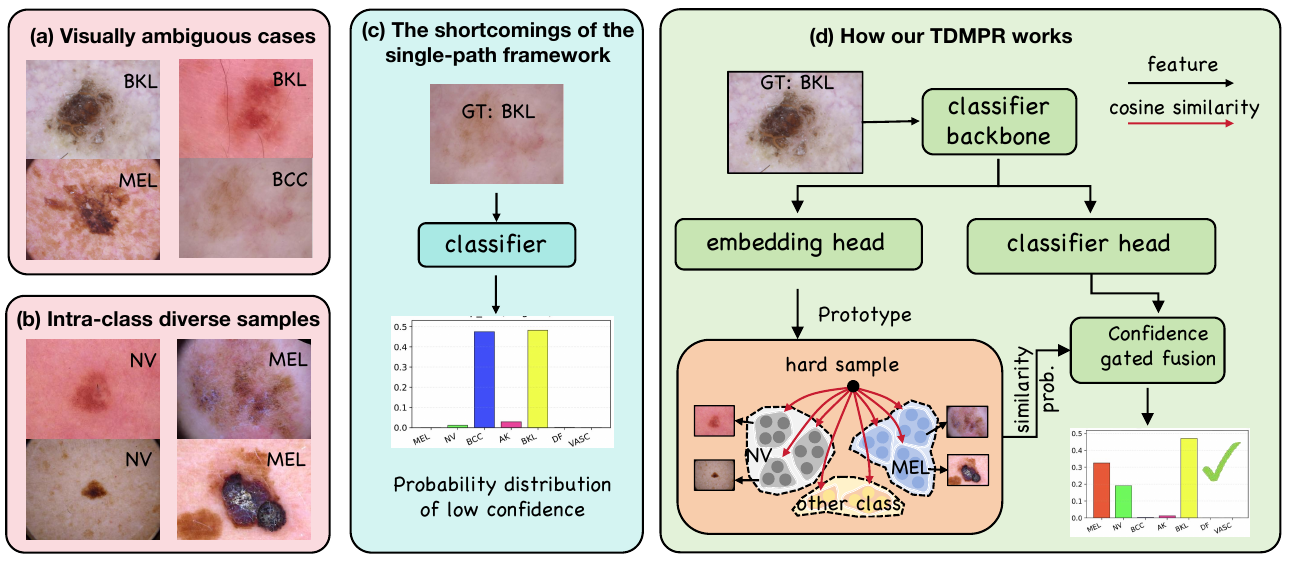}
    \caption{The challenges in fine-grained medical image classification and our method's overview. (a) shows visually ambiguous cases where different categories share similar patterns, leading to classifier uncertainty. (b) highlights intra-class diversity, demonstrating the challenge of handling different appearances within the same category. (c) illustrates the shortcomings of the single-path framework, where predictions become miscalibrated in ambiguous cases. (d) shows how T-DuMpRa integrates a classifier and a similarity pathway, with prototype retrieval and confidence-gated fusion for selective decision-making. This dual-path approach enhances performance on ambiguous samples by combining discriminative classification and reliable similarity-based evidence.}
    \label{fig:fig1}
\end{figure}

Fine-grained medical image classification is a core component of computer-aided diagnosis systems~\cite{zhu2024sfpl,cai2020review}. In dermatology, dermoscopic screening aims to distinguish malignant lesions from benign ones and to refer suspicious cases for further examination~\cite{dinnes2018dermoscopy,alam2025artificial}. A key challenge is visually ambiguous cases~\cite{bresciani2015pitfalls,tang2025mibf}, where different categories share very similar patterns (as shown in Fig.~\ref{fig:fig1}(a)) and the decision depends on subtle cues. Under such ambiguity, purely discriminative classifiers can produce poorly calibrated predictions and struggle distinguishing between highly similar categories. Therefore, practical deployment requires not only high overall accuracy but also reliable performance on visually ambiguous cases~\cite{manhardt2019explaining,nguyen2022trustworthy,van2022explainable}. It also requires decision evidence that clinicians can check, such as retrieving similar reference patterns, because single-path framework (i.e., the framework trained with cross-entropy that directly outputs class probabilities predictions) are hard to trust when visual cues are unclear~\cite{zadeh2020bias,hasani2022trustworthy}.

Modern deep single-path classifiers trained with cross-entropy achieve strong accuracy on standard benchmarks~\cite{valmadre2022hierarchical,kim2022transfer,chen2025review,liu2022acpl}. However, their performance in dermoscopic screening is often limited by visually ambiguous cases. When categories appear similar and discriminative cues are weak, the classifier may output near-equal probabilities for different classes (as shown in Fig.~\ref{fig:fig1}(c)), reflecting uncertainty in its decision. Such uncertainty is particularly undesirable as it can mislead clinical decisions on the most challenging samples. Moreover, real-world medical datasets frequently contain under-represented subtypes and diverse appearances due to variations across patients and imaging conditions~\cite{wen2022characteristics}. A single decision boundary may fit common patterns well but fail on visually ambiguous cases, causing class-balanced metrics to lag behind even when overall accuracy remains high~\cite{li2025unified}. In dermoscopic screening, such probability uncertainty on visually ambiguous cases can directly disrupt triage and biopsy/referral decisions, leading to missed or delayed diagnosis of high-risk lesions while also increasing unnecessary procedures and patient burden~\cite{kittler2002diagnostic,combalia2022validation}. These observations motivate a framework that remains reliable on ambiguous samples and provides checkable evidence to support its predictions.

To address these limitations, we propose T-DuMpRa, a Teacher-guided Dual-path Multi-Prototype Retrieval-augmented framework for fine-grained medical image classification. As illustrated in Fig.~\ref{fig:fig1}(d), T-DuMpRa augments a standard discriminative classifier with a similarity-based retrieval pathway, enabling ambiguous inputs to be supported by reference patterns rather than relying solely on a single decision score. During training, we optimize cross-entropy jointly with supervised contrastive learning to shape a cosine-friendly embedding space, making similarity comparisons more reliable when categories are visually alike. However, the intrinsic diversity of lesion features within the same class poses challenges for similarity retrieval. To address this, we employ an EMA teacher to generate smoother representations and construct a compact multi-prototype memory by clustering the teacher's embeddings. This memory captures the diverse appearance modes within each category while reducing sensitivity to noisy features (as shown in Fig.~\ref{fig:fig1}(d)). At test time, we compute both the classifier's logits and a similarity-based distribution derived from cosine matching against the prototype bank. Critically, we introduce a conservative confidence-gated fusion mechanism: retrieval is activated only when the classifier is uncertain and the similarity evidence is both strong and disagrees with the classifier's prediction; otherwise, the original prediction is preserved. This design selectively targets visually ambiguous cases for correction while avoiding unnecessary changes on easy samples. Moreover, T-DuMpRa is plug-and-play, can be integrated on top of existing classifiers with minimal overhead, and provides checkable similarity evidence to support decisions under ambiguity. Our contributions are four-fold:
\begin{itemize}
    \item \textbf{Teacher-guided dual-path framework.} We introduce T-DuMpRa, a teacher-guided dual-path multi-prototype retrieval-augmented framework that complements standard discriminative classification with a similarity pathway for more reliable decisions on visually ambiguous cases.
    \item \textbf{Stable multi-prototype representation for reliable retrieval.} We learn cosine-compatible embeddings and build a compact multi-prototype bank by clustering EMA-teacher representations, capturing intra-class appearance diversity while reducing sensitivity to noisy or drifting features.
    \item \textbf{Conservative confidence-gated fusion.} We activate retrieval assistance only when the classifier is uncertain and the similarity evidence is decisive and conflicting, preserving predictions on easy samples while enabling targeted correction on ambiguous ones.
    \item \textbf{Plug-and-play with checkable evidence.} The method can be easily integrated into existing classifiers, and it provides prototype-based supporting evidence to help inspect decisions under ambiguity.
\end{itemize}

\section{Related Work}
\label{sec:related}

\subsection{Fine-Grained Medical Image Classification}

Fine-grained medical image classification presents unique challenges due to subtle inter-class differences, high intra-class variance caused by patient heterogeneity, and the frequent occurrence of long-tailed distributions~\cite{liang2025medfilip,chen2024medical,spolaor2024fine,meng2024correlation}. Traditional Convolutional Neural Networks (CNNs) and Vision Transformers (ViTs) often struggle to capture the localized, discriminative features required for this task~\cite{patricio2023explainable,shao2024hybrid,hussain2025effresnet,khan2025recent}. To address this, previous works have heavily relied on attention mechanisms~\cite{cheng2022resganet,ling2023mtanet}, part-based models~\cite{aleem2024test,song2024posture}, and multi-scale feature fusion~\cite{zhu2024lightweight} to force the network to focus on discriminative regions. While these approaches improve overall accuracy, they predominantly rely on purely parametric decision boundaries. Consequently, they remain highly vulnerable to visually ambiguous or "hard" samples that lie near the decision boundaries, often producing incorrect predictions~\cite{kumar2024medical}. In contrast, our method explicitly tackles these hard cases by introducing a dual-path framework that supplements the parametric classifier with non-parametric retrieval evidence.

\subsection{Prototype-Based and Retrieval-Augmented Learning}
Retrieval-augmented models, such as $k$-NN classifiers integrated with deep networks, have shown great promise in improving model interpretability and robustness, particularly in data-scarce or long-tailed scenarios~\cite{zhao2025retrieval,yang2025revisiting,rao2025amd,long2022retrieval}. A prominent branch of this paradigm is prototype-based learning, popularized by Prototypical Networks~\cite{snell2017prototypical}, which classifies samples based on their distance to class representations. Subsequent works have adapted prototypes for fine-grained and medical tasks to provide case-based reasoning~\cite{hu2021semi,huy2025interactive,pellicer2025protomedx,cheng2023prior,sacha2023protoseg}. However, the vast majority of these methods collapse each class into a single mean vector. This over-smoothing destroys the complex intra-class geometry and multi-modality inherent in medical datasets (e.g., variations due to different imaging devices or disease subtypes). Unlike these single-prototype methods, we utilize spherical $k$-means to construct multiple unit-norm prototypes per class. Furthermore, by modeling the prototype posterior as a von Mises–Fisher (vMF) mixture~\cite{hu2025probabilistic,conti2022mitigating,hasnat2017mises}, our approach marginalizes over multiple intra-class modes, explicitly preserving the geometric diversity of the data.
\subsection{Representation Learning and Stable Memory Spaces}
The quality of retrieval-based models is bottlenecked by the discriminative power and stability of the underlying feature space. Supervised Contrastive Learning (SCL)~\cite{khosla2020supervised} has emerged as a powerful tool to pull samples from the same class together while pushing apart different classes, creating a "retrieval-friendly" embedding geometry~\cite{mildenberger2025tale}. However, dynamically updating a memory bank or prototype set during stochastic optimization often leads to embedding drift and representation noise~\cite{liang2025advancing,cao2025few,wen2021toward}. To stabilize feature spaces, Exponential Moving Average (EMA) teachers have been widely utilized in Semi-Supervised Learning (e.g., Mean Teacher)~\cite{tarvainen2017mean} and Self-Supervised Learning (e.g., MoCo)~\cite{he2020momentum}. Inspired by these momentum-based approaches [44-46], we repurpose the EMA teacher architecture to construct our multi-prototype memory~\cite{wang2021tripled}. Because the EMA teacher evolves smoothly, it fundamentally mitigates embedding drift, providing a highly stable reference space that is crucial for reliable similarity matching in fine-grained medical tasks.
\subsection{Uncertainty Estimation and Selective Fusion}
Effectively combining multiple decision streams is a core challenge in ensemble learning and Mixture of Experts (MoE) architectures~\cite{zhou2022mixture,chen2022towards}. Common fusion strategies, such as simple averaging, concatenation, or globally learned attention weights, often suffer from indiscriminate fusion~\cite{li2025deep,nagrani2021attention,han2022multimodal}. That is, the secondary expert may inadvertently degrade the performance of the primary classifier on "easy" samples where the classifier is already highly confident and correct. To prevent this, recent advances in selective classification~\cite{goren2024hierarchical,xu2025dual,geifman2017selective} suggest that interventions should be conditional. Building upon this philosophy, we propose a rigorous confidence-gated fusion mechanism. By simultaneously evaluating classifier uncertainty (via top-1 probability and entropy), prototype reliability, and branch disagreement (via Jensen–Shannon divergence)~\cite{englesson2021generalized,sutter2020multimodal}, our model conservatively activates prototype assistance only when strictly necessary. This selective mechanism mathematically ensures that the risk decomposition confines changes only to the gated subset, protecting baseline performance on confident samples.

\begin{figure}[tb]
    \centering
    \includegraphics[width=\textwidth]{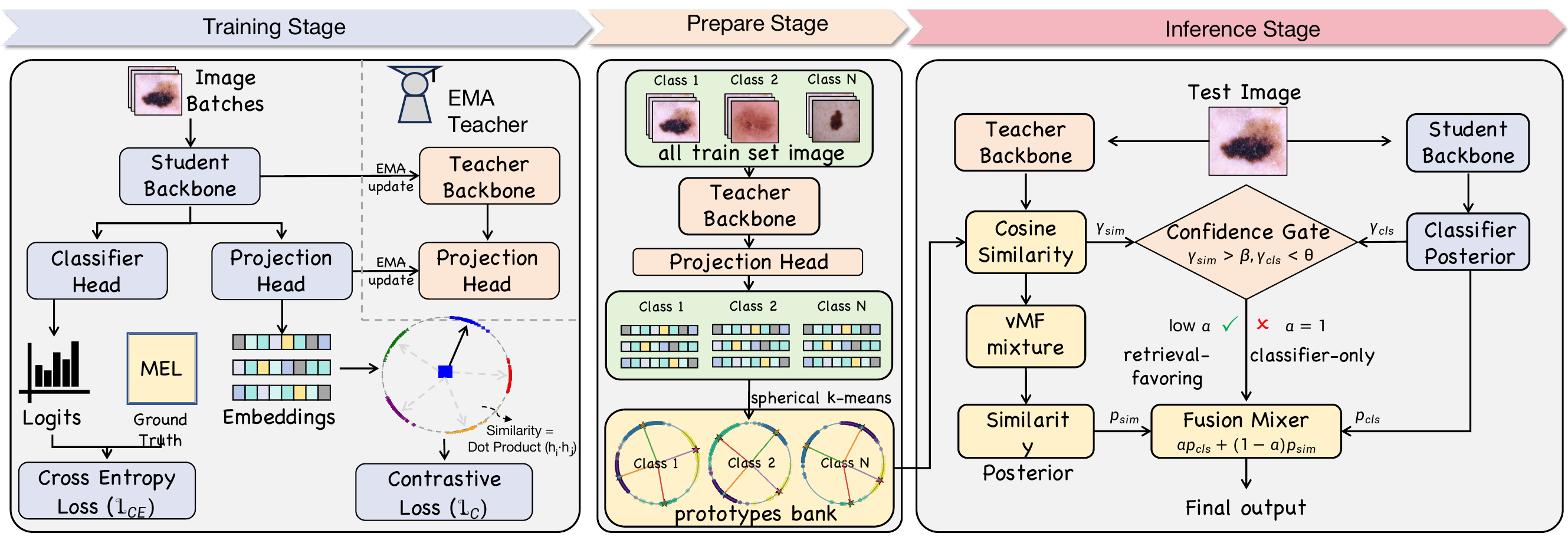}
    \caption{\textbf{The proposed teacher-guided prototype retrieval framework.} 
    \textbf{(a) Joint Training \& Prototype Construction:} The student backbone is optimized via cross-entropy ($\mathcal{L}_{\mathrm{CE}}$) on the classifier head and supervised contrastive loss ($\mathcal{L}_{\mathrm{SCL}}$) on the projection head. An EMA teacher network slowly updates from the student to provide stable representations. After training, multi-modal class prototypes are constructed by clustering teacher embeddings to form a compact prototype bank $\mathcal{P}$. 
    \textbf{(b) Confidence-Gated Dual-Path Inference:} Given a test image, we compute the discriminative posterior $p_{\mathrm{cls}}$ (via student) and the similarity-based posterior $p_{\mathrm{sim}}$ (via teacher embeddings matching against $\mathcal{P}$). A confidence gate $g(x)$ activates retrieval-favoring fusion only when the classifier is uncertain ($\gamma_{\mathrm{cls}} < \theta$) and similarity evidence is reliable, otherwise retaining the classifier's prediction. This design improves robustness on hard cases without compromising easy ones.}
    \label{fig:framework}
\end{figure}

\section{Preliminary}
\label{sec:preliminary}

\subsection{Problem definition.}

We consider the task of fine-grained medical image classification, where we are given a labeled dataset $\mathcal{D} = \{(x_i, y_i)\}_{i=1}^{N}$, with $x_i$ as the input image and $y_i \in \{1, \dots, C\}$ as the class label. The goal is to predict the class label for each image. However, this task is extremely challenging due to the presence of ambiguous hard samples. Effectively addressing these visually ambiguous cases is the key to overcoming performance bottlenecks in the model.

\subsection{Dual-path  framework and its performance upper bound}
The dual-path framework is a feasible solution to this problem which combining two decision paths: (1) Parametric classifierfor direct classification. (2) Prototype retrieval for assisting in uncertain cases. Each path provides a prediction $\hat{y}_e(x) \in \{1, \dots, C\}$, where expert $e \in \{0, 1\}$ corresponds to the classifier or prototype retrieval. This model can be viewed as a two-expert decision system, and it has a provable performance upper bound (as shown in~\ref{sec:appendix:theory:moe}). However, since the conditional risks $r_e(x)$ in formula~\ref{eq:app:bayes_gate} are difficult to predict, a reasonable mechanism is needed to balance the decision results of the two paths in order to achieve better classification performance for visually ambiguous cases.

\section{Method}
\label{sec:method}
We propose a teacher-guided dual-path inference framework for fine-grained medical image classification.
The framework couples a parametric classifier with a non-parametric prototype retrieval branch.
The key is to (i) learn retrieval-friendly embeddings via supervised contrastive learning, (ii) construct stable class prototypes in an EMA teacher space, and (iii) activate retrieval evidence only when it is reliable and the classifier is uncertain. Our method focuses on addressing two issues:
First, to reduce over-confident errors on hard or ambiguous cases, we avoid indiscriminate fusion and instead correct predictions only when the classifier is uncertain and retrieval evidence is strong.
Second, to improve class-wise robustness under long-tailed and visually overlapping categories, we reshape the embedding geometry and explicitly model intra-class multi-modality with multiple prototypes per class.

A formal effectiveness analysis is provided in Appendix~\ref{sec:appendix:theory}, which characterizes (i) the conservative behavior on non-gated samples (Appendix~\ref{sec:appendix:theory:conservative}), (ii) the risk decomposition that confines changes to the gated subset (Eq.~\eqref{eq:app:risk_decomp}), and (iii) sufficient conditions under which gated fusion follows decisive retrieval evidence (Eq.~\eqref{eq:app:alpha_bound}).


\subsection{Teacher-guided representation and prototype memory}
\label{sec:method:teacher_proto}
Prototype quality is sensitive to representation noise, so we construct prototypes in an EMA teacher space to obtain a more stable memory for retrieval.

\paragraph{EMA teacher.}
We maintain a teacher model $(\bar{\theta},\bar{\phi})$ as an exponential moving average of student parameters and use it to extract stable embeddings:
\begin{equation}
\bar{\theta}\leftarrow \mu\bar{\theta}+(1-\mu)\theta,\quad
\bar{\phi}\leftarrow \mu\bar{\phi}+(1-\mu)\phi,\quad
\bar{z}(x)=
\frac{g_{\bar{\phi}}(f_{\bar{\theta}}(x))}
{\|g_{\bar{\phi}}(f_{\bar{\theta}}(x))\|_2}\in\mathbb{S}^{D-1}.
\label{eq:ema_teacher}
\end{equation}
Because the teacher evolves smoothly, its embeddings vary less across updates and are less affected by stochastic optimization.
This reduces embedding drift when building the prototype bank, which is important for fine-grained medical data where subtle appearance variations can otherwise perturb neighborhood structure.

\paragraph{Multi-prototype modeling via spherical $k$-means.}
Medical categories can be multi-modal due to patient heterogeneity, imaging devices, and acquisition conditions.
To preserve this diversity, we represent each class with multiple prototypes instead of a single class mean.
For each class $c$, we collect teacher embeddings $\mathcal{Z}_c=\{\bar{z}(x_i): y_i=c\}$ and construct $K$ unit-norm prototypes $\mathcal{P}_c=\{p_{c,1},\dots,p_{c,K}\}$ with $\|p_{c,k}\|_2=1$.
We solve the spherical clustering objective
\begin{equation}
\min_{\{p_{c,k}\}}
\sum_{\bar{z}\in\mathcal{Z}_c}
\min_{k\in\{1,\dots,K\}}
\left(1-\bar{z}^{\top}p_{c,k}\right)
\quad
\text{s.t.}\quad \|p_{c,k}\|_2=1.
\label{eq:spherical_kmeans_obj}
\end{equation}
Its alternating updates yield a normalized centroid form: if $\mathcal{A}_{c,k}$ is the assigned cluster set, then
\begin{equation}
p_{c,k}
\leftarrow
\mathrm{norm}\!\left(\sum_{\bar{z}\in\mathcal{A}_{c,k}} \bar{z}\right),
\label{eq:spherical_kmeans_update}
\end{equation}
where $\mathrm{norm}(v)=v/\|v\|_2$.
All prototypes form a compact memory bank $\mathcal{P}\in\mathbb{R}^{C\times K\times D}$.
Multiple prototypes provide localized anchors for similarity matching and avoid over-smoothing intra-class modes, which helps reduce confusion when inter-class boundaries are subtle and classes overlap visually.

\subsection{Prototype posterior as a von Mises--Fisher mixture}
\label{sec:method:proto_posterior}
Given a test image $x$, we compute $\bar{z}(x)$ and match it against $\mathcal{P}$.
We denote cosine similarity as $s_{c,k}(x)=\bar{z}(x)^{\top}p_{c,k}\in[-1,1]$.

A principled view is to treat each prototype as a component direction of a von Mises--Fisher (vMF) distribution on $\mathbb{S}^{D-1}$.
Under a mixture of vMF components, the unnormalized class score is
\begin{equation}
q_c(x)
=
\log
\sum_{k=1}^{K}
\exp\!\left(\kappa\, s_{c,k}(x)\right),
\label{eq:vmf_mixture_score}
\end{equation}
where $\kappa>0$ is a concentration parameter.
The log-sum-exp aggregation provides a smooth approximation to max pooling while still emphasizing strong matches.
It can be interpreted as marginalizing over multiple intra-class modes, which improves robustness when class evidence is distributed across several prototypes rather than dominated by a single nearest center.

We then define the prototype-based posterior
\begin{equation}
p_{\mathrm{sim}}(y=c\mid x)
=
\frac{\exp\!\left(q_c(x)/\tau_{\mathrm{sim}}\right)}
{\sum_{j=1}^{C}\exp\!\left(q_j(x)/\tau_{\mathrm{sim}}\right)}.
\label{eq:sim_posterior}
\end{equation}
This yields a similarity posterior that can be fused with the classifier posterior in a probabilistic and interpretable manner.
In Appendix~\ref{sec:appendix:theory:follow_sim}, we further show that decisive similarity margins, together with a low-confidence classifier, provide a sufficient condition for the fused decision to follow retrieval evidence (Eq.~\eqref{eq:app:alpha_bound}).

\subsection{Confidence-gated fusion}
\label{sec:method:gated_fusion}

\paragraph{Confidence and disagreement signals.} We design a gate using uncertainty and reliability signals so that retrieval evidence is used only when it is likely to correct the classifier, rather than perturbing already-correct predictions~\cite{yan2026confidence}. We use the top-1 probability as classifier uncertainty and measure distributional uncertainty using entropy~\cite{kendall2017uncertainties}:
\begin{equation}
\gamma_{\mathrm{cls}}(x)=\max_{c}p_{\mathrm{cls}}(y=c\mid x).
\label{eq:gamma_cls}
\end{equation}

\begin{equation}
H_{\mathrm{cls}}(x)
=
-\sum_{c=1}^{C} p_{\mathrm{cls}}(y=c\mid x)\log p_{\mathrm{cls}}(y=c\mid x).
\label{eq:entropy_cls}
\end{equation}

\paragraph{Prototype reliability.}
We define $\gamma_{\mathrm{sim}}(x)=\max_{c}p_{\mathrm{sim}}(y=c\mid x)$ and margin $\Delta_{\mathrm{sim}}(x)=p_{\mathrm{sim}}^{(1)}(x)-p_{\mathrm{sim}}^{(2)}(x)$, where $p_{\mathrm{sim}}^{(1)}$ and $p_{\mathrm{sim}}^{(2)}$ are the largest and second-largest values.
High $\gamma_{\mathrm{sim}}$ together with a large margin indicates that retrieval evidence is decisive rather than ambiguous, which helps filter out unstable similarity matches.

\paragraph{Branch disagreement.}
Beyond label disagreement, we quantify distributional mismatch via Jensen--Shannon divergence
\begin{equation}
D_{\mathrm{JS}}(x)
=
\mathrm{JS}\!\left(p_{\mathrm{cls}}(\cdot\mid x)\,\|\,p_{\mathrm{sim}}(\cdot\mid x)\right),
\label{eq:js_div}
\end{equation}
which is symmetric and bounded.
A large divergence indicates a meaningful conflict between discriminative and retrieval-based explanations, which is precisely the scenario where selective correction is most valuable.
From the mixture-of-experts view, these signals serve as observable surrogates for comparing expert risks; Appendix~\ref{sec:appendix:theory:surrogate} formalizes this connection via a local confidence approximation (Eq.~\eqref{eq:app:local_calib}) and a sufficient condition for Bayes-consistent selection (Eq.~\eqref{eq:app:gap_condition}).


We activate prototype assistance only when the classifier is uncertain and the prototype evidence is reliable.
Let $\hat{y}_{\mathrm{cls}}=\arg\max_{c}p_{\mathrm{cls}}(y=c\mid x)$ and $\hat{y}_{\mathrm{sim}}=\arg\max_{c}p_{\mathrm{sim}}(y=c\mid x)$.
We define a binary gate
\begin{equation}
g(x)
=
\mathbb{I}\!\left[\gamma_{\mathrm{cls}}(x)<\theta\right]\cdot
\mathbb{I}\!\left[\gamma_{\mathrm{sim}}(x)>\beta\right]\cdot
\mathbb{I}\!\left[\Delta_{\mathrm{sim}}(x)>m_{\mathrm{sim}}\right]\cdot
\mathbb{I}\!\left[D_{\mathrm{JS}}(x)>\delta\right]\cdot
\mathbb{I}\!\left[\hat{y}_{\mathrm{cls}}\neq \hat{y}_{\mathrm{sim}}\right],
\label{eq:gate_full}
\end{equation}
where $\theta$ controls when the classifier is considered uncertain, $(\beta,m_{\mathrm{sim}})$ enforce prototype reliability, and $\delta$ avoids activation on near-identical posteriors.
When $g(x)=1$, we apply retrieval-favoring fusion
\begin{equation}
p_{\mathrm{fuse}}(\cdot\mid x)
=
\alpha_{\mathrm{low}}\,p_{\mathrm{cls}}(\cdot\mid x)
+
(1-\alpha_{\mathrm{low}})\,p_{\mathrm{sim}}(\cdot\mid x).
\label{eq:fuse_when_on}
\end{equation}
Otherwise, we keep the classifier prediction unchanged.
The final posterior is
\begin{equation}
p(\cdot\mid x)
=
(1-g(x))\,p_{\mathrm{cls}}(\cdot\mid x)
+
g(x)\,p_{\mathrm{fuse}}(\cdot\mid x).
\label{eq:final_posterior} 
\end{equation}
This conservative policy preserves the classifier output on confident samples, while enabling targeted corrections on hard cases where the classifier is less trustworthy and retrieval evidence is both strong and conflicting.
Appendix~\ref{sec:appendix:theory:conservative} formalizes the invariance of non-gated predictions, and Eq.~\eqref{eq:app:risk_decomp} shows that any risk change is concentrated on the gated subset.
Moreover, Appendix~\ref{sec:appendix:theory:follow_sim} provides a sufficient condition under which the fused decision follows the similarity expert when the gate triggers (Eq.~\eqref{eq:app:alpha_bound}), which directly supports the intended correction behavior.

\subsection{Training objective: discriminative learning with supervised contrastive geometry}
\label{sec:method:train}
We optimize the backbone with a joint objective that aligns discriminative classification with metric structure learning.

\paragraph{Cross-entropy loss.}
For each sample $(x,y)$, we use $\mathcal{L}_{\mathrm{CE}}(x,y)=-\log p_{\mathrm{cls}}(y\mid x)$.

\paragraph{Supervised contrastive loss.}
For each image $x_i$, we generate two augmented views $\{x_i^{(1)},x_i^{(2)}\}$.
Let $\mathcal{I}$ index all views in the batch (size $2B$), and let $z_i$ denote the embedding of view $i$.
For an anchor $i\in\mathcal{I}$, its positive set is $\mathcal{P}(i)=\{p\in\mathcal{I}\setminus\{i\}: y_p=y_i\}$.
The supervised contrastive loss (SCL) is
\begin{equation}
\mathcal{L}_{\mathrm{SCL}}
=
\sum_{i\in\mathcal{I}}
\frac{-1}{|\mathcal{P}(i)|}
\sum_{p\in\mathcal{P}(i)}
\log
\frac{\exp\!\left(z_i^{\top}z_p/\tau\right)}
{\sum_{a\in\mathcal{I}\setminus\{i\}}
\exp\!\left(z_i^{\top}z_a/\tau\right)}.
\label{eq:supcon_full}
\end{equation}

Unlike cross-entropy, which mainly adjusts decision boundaries via logits, $\mathcal{L}_{\mathrm{SCL}}$ explicitly shapes the representation geometry by tightening intra-class neighborhoods and separating different classes on the hypersphere.
This property is especially beneficial under long-tailed and overlapping categories~\cite{li2022targeted} because it makes cosine similarity more faithful to semantic proximity and reduces brittle nearest-neighbor behavior in poorly structured spaces~\cite{sharma2024confidence}.
From the gating perspective, a better-structured similarity posterior also reduces the probability of selecting the worse expert; Appendix~\ref{sec:appendix:theory:surrogate} connects this goal to the gate regret bound (Eq.~\eqref{eq:app:regret_bound}).

\paragraph{Overall training loss.}
The final training objective is
\begin{equation}
\mathcal{L}
=
\mathbb{E}_{(x,y)\sim \mathcal{D}}
\left[
\mathcal{L}_{\mathrm{CE}}(x,y)
+
\lambda\,\mathcal{L}_{\mathrm{SCL}}(x,y)
\right].
\label{eq:train_total}
\end{equation}

\subsection{Complexity and overhead}
\label{sec:method:complexity}
The summary of our method is shown in the Algorithm~\ref{alg:method}.The prototype bank stores $C\times K$ vectors in $\mathbb{R}^{D}$, which is lightweight.
For each test sample, the extra cost is a matrix multiplication between $\bar{z}(x)$ and the prototype bank, i.e., $\mathcal{O}(CKD)$ operations, which is typically negligible compared with a backbone forward pass.
The overhead is controllable through $K$ and $D$.
Importantly, the framework is plug-and-play: it requires no backbone modification and introduces only a compact prototype bank and lightweight similarity computation at inference.

\section{Experiments}
\label{sec:exp}

\subsection{Dataset and Protocol}
\label{sec:exp:data}
We evaluate on \textbf{HAM10000}, a dermoscopic lesion classification benchmark with $C{=}7$ categories and a long-tailed label distribution.
We follow the official train/test split provided with the dataset.
From the training split, we further construct a \textbf{stratified validation} subset (10\% of the training samples) for selecting inference-time hyperparameters that affect fusion (e.g., $\theta$, $\beta$, $m_{\text{sim}}$, $\tau_{\text{sim}}$).
After hyperparameters are fixed, we optionally retrain the model on \textbf{train+val} and report the final performance on the held-out test set.
Unless otherwise specified, we report the mean$\pm$std over three random seeds.

\subsection{Evaluation Metrics}
\label{sec:exp:metrics}
We report: \textbf{Accuracy} (Acc), \textbf{Macro-F1}, and \textbf{Balanced Accuracy} (BalAcc, mean recall across classes).
Since medical datasets often exhibit class imbalance, BalAcc is treated as a primary indicator of robustness to long-tailed distributions.
We additionally report \textbf{macro-AUROC} (mAUC) computed in a one-vs-rest manner and averaged over classes.
To support our confidence-gated design, we include \textbf{Expected Calibration Error} (ECE, lower is better), computed with $M{=}15$ equal-width confidence bins~\cite{tsuneki2022deep}.

\begin{table}[t]
\centering
\caption{Compare experiment results across backbones on HAM10000.}
\label{tab:main_backbone}

\setlength{\tabcolsep}{3.8pt}
\renewcommand{\arraystretch}{1.15}

\resizebox{\textwidth}{!}{%
\begin{tabular}{lcccccc}
\toprule
Method
& Acc$\uparrow$
& Macro-F1$\uparrow$
& BalAcc$\uparrow$
& ECE$\downarrow$
& mAUC$\uparrow$
& Delta Acc$\uparrow$ \\
\midrule

ConvNeXt~\cite{liu2022convnet}
& 0.8800$\pm$0.0032 & 0.7991$\pm$0.0051 & 0.7934$\pm$0.0048 & 0.0862$\pm$0.0035 & 0.9773$\pm$0.0012 & - \\
\rowcolor{gray!12}
\textbf{w/ T-DuMpRa}
& 0.8951$\pm$0.0028 & 0.8090$\pm$0.0045 & 0.8012$\pm$0.0042 & 0.0834$\pm$0.0031 & 0.9793$\pm$0.0010 & \textcolor{red}{+1.72\%} \\
\midrule

Efficientnet~\cite{tan2019efficientnet}
& 0.7675$\pm$0.0085 & 0.5702$\pm$0.0102 & 0.5516$\pm$0.0120 & 0.1298$\pm$0.0055 & 0.9262$\pm$0.0018 & - \\
\rowcolor{gray!12}
\textbf{w/ T-DuMpRa}
& 0.7845$\pm$0.0072 & 0.5999$\pm$0.0095 & 0.5772$\pm$0.0105 & 0.1030$\pm$0.0048 & 0.9266$\pm$0.0016 & \textcolor{red}{+2.21\%} \\
\midrule

ResNet~\cite{He_2016_CVPR}
& 0.8165$\pm$0.0058 & 0.6422$\pm$0.0080 & 0.5896$\pm$0.0090 & 0.1314$\pm$0.0050 & 0.9494$\pm$0.0015 & - \\
\rowcolor{gray!12}
\textbf{w/ T-DuMpRa}
& 0.8220$\pm$0.0052 & 0.6659$\pm$0.0075 & 0.6304$\pm$0.0085 & 0.1269$\pm$0.0045 & 0.9545$\pm$0.0013 & \textcolor{red}{+0.67\%} \\
\midrule

ViT~\cite{dosovitskiy2020image}
& 0.8265$\pm$0.0050 & 0.6903$\pm$0.0065 & 0.6614$\pm$0.0070 & 0.1248$\pm$0.0042 & 0.9515$\pm$0.0014 & - \\
\rowcolor{gray!12}
\textbf{w/ T-DuMpRa}
& 0.8480$\pm$0.0042 & 0.7182$\pm$0.0060 & 0.6635$\pm$0.0065 & 0.1188$\pm$0.0040 & 0.9525$\pm$0.0012 & \textcolor{red}{+2.61\%} \\
\midrule

SwinViT~\cite{liu2021swin}
& 0.8960$\pm$0.0025 & 0.8095$\pm$0.0040 & 0.7860$\pm$0.0045 & 0.0781$\pm$0.0030 & 0.9787$\pm$0.0010 & - \\
\rowcolor{gray!12}
\textbf{w/ T-DuMpRa}
& 0.9030$\pm$0.0022 & 0.8229$\pm$0.0035 & 0.7992$\pm$0.0040 & 0.0766$\pm$0.0025 & 0.9800$\pm$0.0008 & \textcolor{red}{+0.78\%} \\

\bottomrule
\end{tabular}%
}
\end{table}

\subsection{Implementation Details}
\label{sec:exp:impl}
\paragraph{Backbones.}
We verify plug-and-play behavior across multiple backbones:
\textbf{ResNet-101}~\cite{He_2016_CVPR}, \textbf{ConvNeXt-Tiny}~\cite{liu2022convnet}, \textbf{Efficientnet-B0}~\cite{tan2019efficientnet}, \textbf{ViT-B}~\cite{dosovitskiy2020image} and \textbf{SwinVit-B}~\cite{liu2021swin}.
All backbones are initialized with ImageNet pretraining.

\paragraph{Training.}
Images are resized to $224\times224$.
We apply standard augmentation for two-view supervised contrastive learning (random resized crop, horizontal flip, color jitter, random grayscale).
We optimize with AdamW (lr $=1\mathrm{e}{-4}$, weight decay $=1\mathrm{e}{-4}$), batch size $=64$, for 20 epochs.
For $\mathcal{L}_{\mathrm{SCL}}$, we use temperature $\tau{=}0.07$ and weight $\lambda{=}0.03$ (kept \textbf{fixed across backbones} unless stated).
We maintain an EMA teacher with momentum $m{=}0.999$ and build the prototype bank in the teacher embedding space using the deterministic transform.

\paragraph{Inference and Fusion.}
We compute the classifier posterior $p_{\text{cls}}$ and prototype similarity posterior $p_{\text{sim}}$ (cosine similarity + softmax with $\tau_{\text{sim}}$).
We use \textbf{confidence-gated fusion} consistent with Eq.~\eqref{eq:gamma_cls}--\eqref{eq:final_posterior}. 
Unless otherwise noted, we set $\alpha_{\text{low}}{=}0.9$.
We select $\theta$ from $\{0.5,0.6,0.7,0.8,0.9\}$ on the validation split, and tune $(\beta,m_{\text{sim}},\tau_{\text{sim}})$ on the same validation split, then keep them fixed for test reporting.

\subsection{Main Results Across Backbones}
\label{sec:exp:main}
The experimental results on the HAM10000 and ISIC2019 datasets demonstrate that incorporating our proposed dual-path decision framework leads to significant performance improvements across all backbone networks. Specifically, both high-performance models like ConvNeXt and ViT, as well as basic models like EfficientNet, show clear enhancements in key metrics such as accuracy, macro-F1, balanced accuracy, and mAUC. On the HAM10000 dataset, ViT and ConvNeXt achieve notable accuracy gains of 2.61\% and 1.72\%, respectively, while also improving macro-F1 and mAUC. Similarly, on ISIC2019, ConvNeXt and ViT show accuracy improvements of 2.69\% and 1.72\%, with additional enhancements in balanced accuracy and mAUC. These results highlight that our framework not only boosts overall classification accuracy but also improves model calibration (lower ECE) and robustness (higher macro-F1 and mAUC), particularly in handling complex and diverse datasets. Overall, our method demonstrates its broad applicability and superior performance in optimizing multiple evaluation metrics across different backbone networks.

\begin{table}[t]
\centering
\caption{Compare experiment results across backbones on ISIC2019.}
\label{tab:main_backbone}

\setlength{\tabcolsep}{3.8pt}
\renewcommand{\arraystretch}{1.15}

\resizebox{\textwidth}{!}{%
\begin{tabular}{lcccccc}
\toprule
Method
& Acc$\uparrow$
& Macro-F1$\uparrow$
& BalAcc$\uparrow$
& ECE$\downarrow$
& mAUC$\uparrow$
& Delta Acc$\uparrow$ \\
\midrule

ConvNeXt~\cite{liu2022convnet}
& 0.7036$\pm$0.0032 & 0.5536$\pm$0.0051 & 0.5127$\pm$0.0048 & 0.2316$\pm$0.0035 & 0.9126$\pm$0.0012 & - \\
\rowcolor{gray!12}
\textbf{w/ T-DuMpRa}
& 0.7226$\pm$0.0028 & 0.5916$\pm$0.0045 & 0.5639$\pm$0.0042 & 0.2136$\pm$0.0031 & 0.9166$\pm$0.0010 & \textcolor{red}{+2.69\%} \\
\midrule

Efficientnet~\cite{tan2019efficientnet}
& 0.6188$\pm$0.0035 & 0.4409$\pm$0.0041 & 0.4169$\pm$0.0037 & 0.2605$\pm$0.0043 & 0.8682$\pm$0.0024 & - \\
\rowcolor{gray!12}
\textbf{w/ T-DuMpRa}
& 0.6247$\pm$0.0032 & 0.4547$\pm$0.0046 & 0.4298$\pm$0.0040 & 0.2490$\pm$0.0038 & 0.8675$\pm$0.0021 & \textcolor{red}{+0.96\%} \\
\midrule

ResNet~\cite{He_2016_CVPR}
& 0.6990$\pm$0.0043 & 0.5474$\pm$0.0060 & 0.5232$\pm$0.0057 & 0.2312$\pm$0.0032 & 0.9094$\pm$0.0021 & - \\
\rowcolor{gray!12}
\textbf{w/ T-DuMpRa}
& 0.7021$\pm$0.0041 & 0.5558$\pm$0.0055 & 0.5288$\pm$0.0051 & 0.2303$\pm$0.0028 & 0.9146$\pm$0.0019 & \textcolor{red}{+0.44\%} \\
\midrule

ViT~\cite{dosovitskiy2020image}
& 0.6307$\pm$0.0029 & 0.4497$\pm$0.0036 & 0.4223$\pm$0.0032 & 0.2656$\pm$0.0038 & 0.8479$\pm$0.0019 & - \\
\rowcolor{gray!12}
\textbf{w/ T-DuMpRa}
& 0.6415$\pm$0.0031 & 0.4526$\pm$0.0040 & 0.4261$\pm$0.0036 & 0.2639$\pm$0.0037 & 0.8604$\pm$0.0018 & \textcolor{red}{+1.72\%} \\
\midrule

SwinViT~\cite{liu2021swin}
& 0.7155$\pm$0.0025 & 0.5843$\pm$0.0031 & 0.5410$\pm$0.0028 & 0.2120$\pm$0.0031 & 0.9087$\pm$0.0015 & - \\
\rowcolor{gray!12}
\textbf{w/ T-DuMpRa}
& 0.7244$\pm$0.0023 & 0.5874$\pm$0.0032 & 0.5599$\pm$0.0030 & 0.2141$\pm$0.0026 & 0.9030$\pm$0.0014 & \textcolor{red}{+1.25\%} \\

\bottomrule
\end{tabular}%
}
\end{table}

\subsection{Ablation Study}
\label{sec:exp:ablation}
We carefully designed extensive ablation studies to demonstrate the effectiveness of our method, and all experiments were conducted on the ViT-B model. The ablation study provides insights into the contributions of each component in our method. First, comparing row \textcircled{1} and row \textcircled{2}, we see that adding the prototype and gated mechanism in the student space results in minimal improvement (Acc: 0.8870 vs. 0.8880). This shows that without $\mathcal{L}_{\mathrm{SCL}}$, the embedding space is not optimized effectively, making the prototype retrieval less effective.
Second, row \textcircled{1} and row \textcircled{3} highlight the impact of $\mathcal{L}{\mathrm{SCL}}$, which boosts accuracy (0.8950 vs. 0.8870) and improves macro-F1 (0.8177 vs. 0.8037). $\mathcal{L}_{\mathrm{SCL}}$ optimizes the embedding space, making it more discriminative, which enhances both the backbone classifier and prototype retrieval performance.
Third, comparing rows \textcircled{3}, \textcircled{4}, and \textcircled{5}, EMA-based teacher prototypes (row \textcircled{5}) lead to improved accuracy (0.8970) and a lower ECE (0.0582). The EMA technique stabilizes the prototypes, reducing noise and improving retrieval reliability.
Finally, row \textcircled{6} with the confidence-gated fusion strategy achieves the highest accuracy (0.9020) and macro-F1 (0.8259). The gated fusion selectively integrates prototype information when the classifier is uncertain, improving performance by using reliable retrieval data only when necessary.
In summary, each component—$\mathcal{L}_{\mathrm{SCL}}$, EMA-based prototypes, and confidence-gated fusion—contributes significantly to the model's performance, enhancing accuracy, F1-score, and calibration.

\begin{table}[t]
\centering
\small
\setlength{\tabcolsep}{3.8pt}
\renewcommand{\arraystretch}{1.15}
\caption{Ablation studies with ViT-B on HAM10000 Dataset.}
\label{tab:ablation}
\resizebox{\textwidth}{!}{
\begin{tabular}{c lcccc}
\toprule
\# & Variant & Acc$\uparrow$ & Macro-F1$\uparrow$ & BalAcc$\uparrow$ & ECE$\downarrow$ \\
\midrule
\textcircled{1} & SwinViT-B  & 0.8870 & 0.8037 & 0.7807 & 0.0764  \\
\textcircled{2} & SwinViT-B + Proto+Gated & 0.8880 & 0.8037 & 0.7807 & 0.0782 \\
\textcircled{3} & SwinViT-B+$\mathcal{L}_{\mathrm{SCL}}$ & 0.8950 & 0.8177 & 0.7994 & 0.0665 \\
\textcircled{4} & SwinViT-B+$\mathcal{L}_{\mathrm{SCL}}$ + Proto + Fixed fusion & 0.8950 & 0.8180 & 0.8008 & 0.0643 \\
\textcircled{5} & SwinViT-B+$\mathcal{L}_{\mathrm{SCL}}$ + EMA + Proto + Fixed fusion & 0.8970 & 0.8194 & 0.8018 & \textbf{0.0582 } \\
\hline
\rowcolor{gray!12}
\textcircled{6} & \textbf{SwinViT-B+$\mathcal{L}_{\mathrm{SCL}}$ +  EMA + Proto + Gated} & \textbf{0.9020} & \textbf{0.8259} &\textbf{ 0.8088 }&  0.0615\\
\bottomrule
\end{tabular}
}
\end{table}

\subsection{Confidence Gating Analysis}
\label{sec:exp:gating}
\begin{figure}[!t]
    \centering
    \begin{subfigure}{0.49\linewidth}
        \centering
        \includegraphics[width=\linewidth]{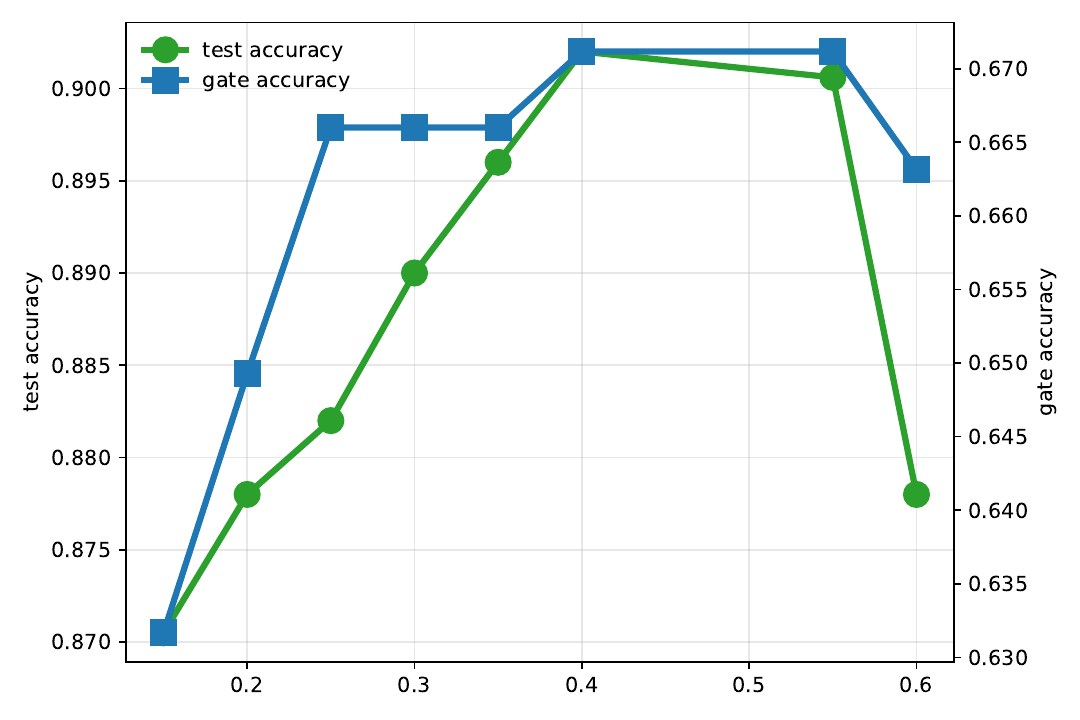}
        \caption{\textbf{Ablation study of $\theta$.}}
    \end{subfigure}
    \hfill
    \begin{subfigure}{0.49\linewidth}
        \centering
        \includegraphics[width=\linewidth]{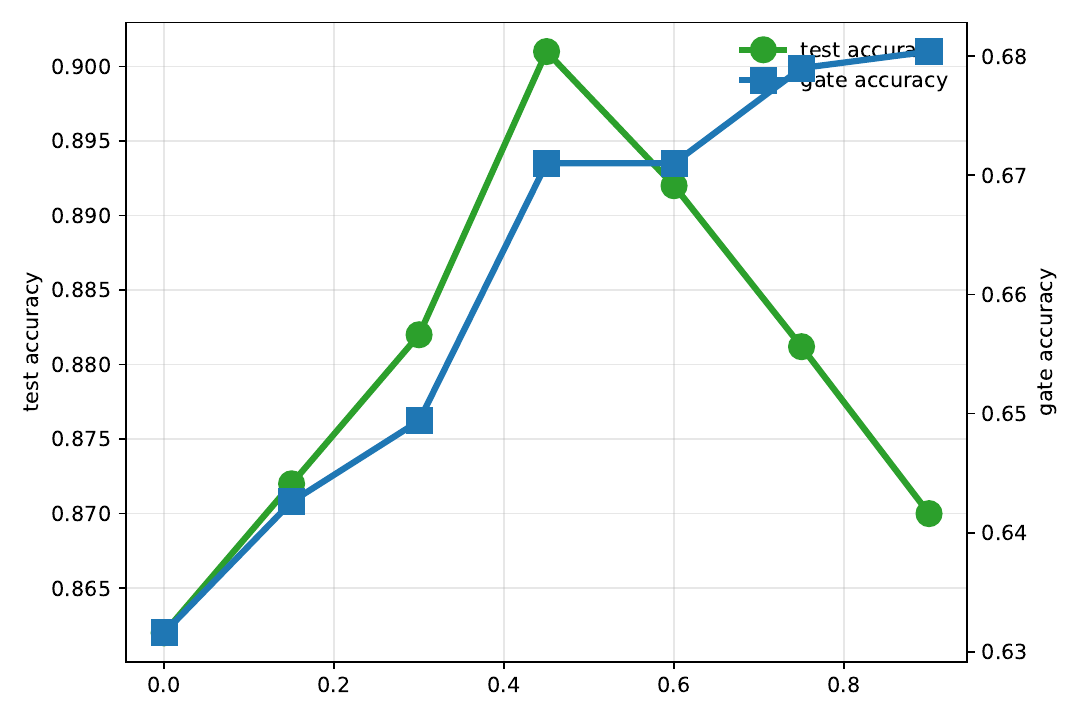}
        \caption{\textbf{Ablation study of $\beta$.}}
    \end{subfigure}
    \caption{\textbf{Results of ablation experiments for classifier confidence threshold for gating $\theta$} and similarity reliability thresholds $\beta$.}
    \label{fig:abl_studies_theta_alpha}
\end{figure}

To verify the effectiveness of gating, we conducted the experiment with SwinViT-B model and the result are shown in Fig.~\ref{fig:abl_studies_theta_alpha}. By changing the values of the classifier confidence threshold for gating, $\theta$, and the similarity reliability threshold, $\beta$, we observed how accuracy changes with these adjustments. We found that as both $\theta$ and $\beta$ increased, the gating accuracy improved because the conditions for gating became stricter. However, due to the reduced throughput after tightening the gating criteria, the overall accuracy first increased and then decreased. This trend aligns with the derivations in the Method section.

\begin{figure}[t]
\centering
\includegraphics[width=\linewidth]{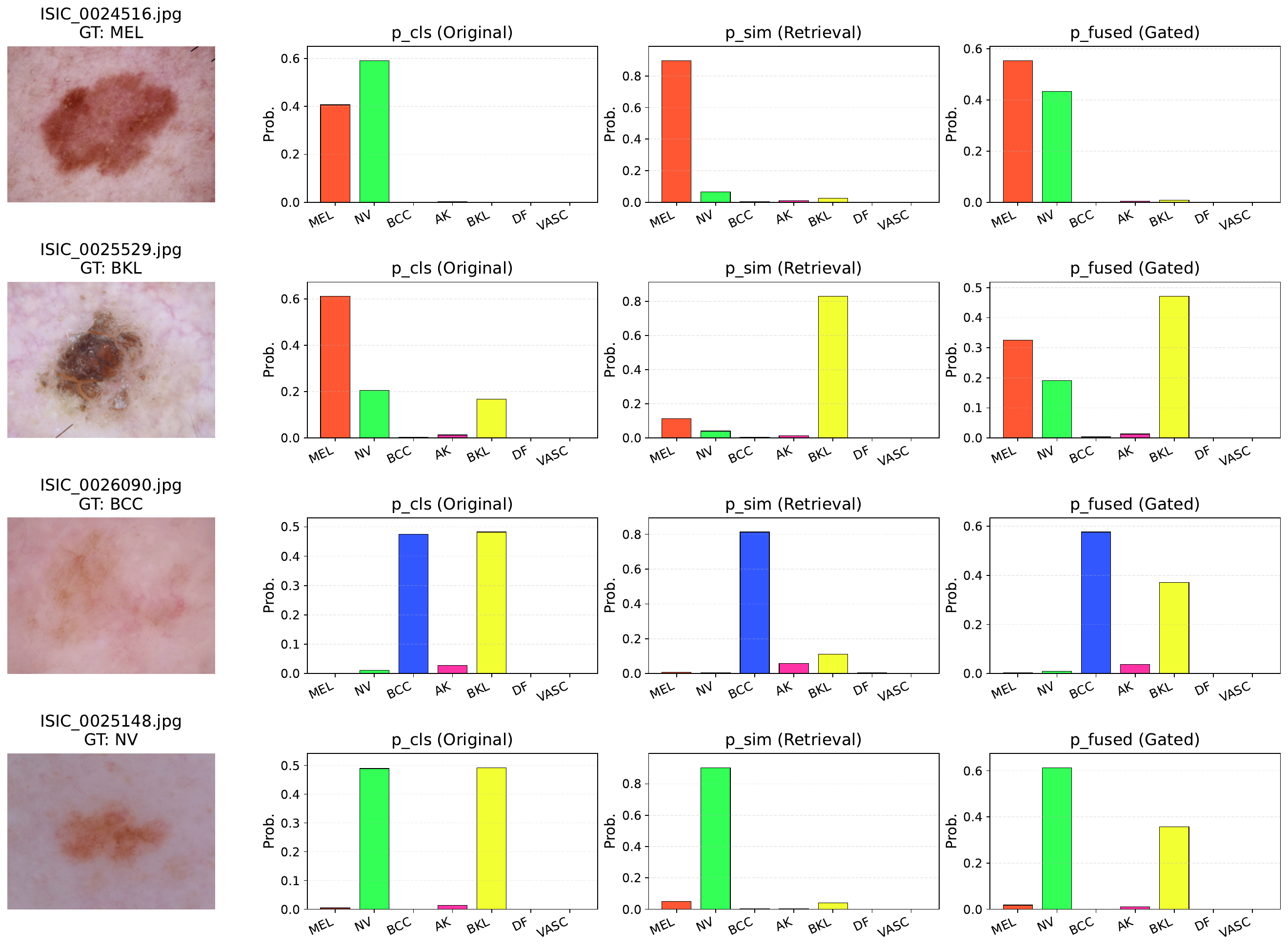}
\caption{Qualitative examples visualization. We visualized the results on HAM using the experimentally optimal hyperparameter setting with the ViT-B model. In this evaluation, we randomly selected four samples for analysis.}
\label{fig:visual}
\end{figure}

\subsection{Visualization analysis.}
\label{sec:exp:qual}
The qualitative examples in Fig.~\ref{fig:visual} demonstrate the effectiveness of our method on low-confidence samples. For instance, in \textbf{ISIC\_0024516} (GT: MEL), the classifier initially predicts MEL with high confidence (0.4), but also assigns higher probabilities to NV (0.6). After incorporating prototype retrieval, the fused posterior boosts the MEL class (with confidence of 0.9), correcting the classifier's uncertainty. Similarly, in \textbf{ISIC\_0026090} (GT: BCC), the classifier is confused (0.47 BCC v.s. 0.48 BKL) and misclassifies the sample as BKL, but prototype retrieval correctly identifies BCC with a high similarity score. The fused prediction increases the BCC confidence, improving the overall accuracy. This shows that our gating mechanism effectively incorporates prototype retrieval when the classifier is uncertain, leading to more reliable predictions, as seen in the improved fused probabilities.

\section{Conclusion}
We propose a dual-path framework for fine-grained medical image diagnosis that integrates discriminative classification with prototype-based similarity reasoning. The method leverages cross-entropy and contrastive learning to structure the embedding space, while an EMA teacher generates stable representations and class prototypes form a reliable memory. During inference, similarity assistance is activated for uncertain samples via confidence-gated fusion. This design enhances robustness on ambiguous cases without sacrificing performance on easier ones. Experiments on HAM10000 show improvements in balanced accuracy. Future work will focus on adaptive prototypes, lesion-aware embeddings, and calibration-aware gating for uncertainty control.

%
%
\bibliographystyle{splncs04}
\bibliography{main}

\newpage
\appendix

\section{Effectiveness Analysis of Confidence-Gated Prototype Retrieval}
\label{sec:appendix:theory}

This appendix provides a theoretical justification for the proposed confidence-gated dual-path inference.
We analyze the method from two complementary perspectives.
First, we give sufficient conditions under which the gated fusion preserves confident classifier predictions and reduces expected risk on the gated subset.
Second, we cast our approach as a two-expert (parametric vs.\ non-parametric) decision system and derive the Bayes-optimal gating rule.
We then show how our practical gate serves as a realizable surrogate of this optimal selector using confidence, margin, and disagreement signals.

\subsection{Setup and notation}
\label{sec:appendix:theory:setup}
Let $x$ be a test sample with ground-truth label $y\in\{1,\dots,C\}$.
Denote the classifier posterior by $p_{\mathrm{cls}}(\cdot\mid x)$ and the prototype posterior by $p_{\mathrm{sim}}(\cdot\mid x)$.
Let
\[
\hat{y}_{\mathrm{cls}}(x)=\arg\max_c p_{\mathrm{cls}}(c\mid x),\qquad
\hat{y}_{\mathrm{sim}}(x)=\arg\max_c p_{\mathrm{sim}}(c\mid x).
\]
The confidence-gated prediction follows Sec.~\ref{sec:method:gated_fusion}.
Define the gate $g(x)\in\{0,1\}$ and the fused posterior
\[
p_{\mathrm{fuse}}(\cdot\mid x)=\alpha_{\mathrm{low}}\,p_{\mathrm{cls}}(\cdot\mid x)+(1-\alpha_{\mathrm{low}})\,p_{\mathrm{sim}}(\cdot\mid x),
\]
and the final posterior
\begin{equation}
p(\cdot\mid x)=(1-g(x))\,p_{\mathrm{cls}}(\cdot\mid x)+g(x)\,p_{\mathrm{fuse}}(\cdot\mid x),
\label{eq:app:final_posterior_def}
\end{equation}
with final prediction $\hat{y}(x)=\arg\max_c p(c\mid x)$.
We analyze the expected $0$-$1$ risk:
\begin{equation}
\mathcal{R}(\hat{y}) \;=\; \mathbb{E}\big[\mathbb{I}[\hat{y}(x)\neq y]\big].
\label{eq:app:risk_def}
\end{equation}

\subsection{Conservative property: invariance on the non-gated set}
\label{sec:appendix:theory:conservative}

\paragraph{Proposition A.1 (Invariance when $g(x)=0$).}
For any sample $x$ with $g(x)=0$, the final posterior equals the classifier posterior and thus the final decision equals the classifier decision:
\[
p(\cdot\mid x)=p_{\mathrm{cls}}(\cdot\mid x),\qquad
\hat{y}(x)=\hat{y}_{\mathrm{cls}}(x).
\]

\paragraph{Proof.}
If $g(x)=0$, Eq.~\eqref{eq:app:final_posterior_def} gives $p(\cdot\mid x)=p_{\mathrm{cls}}(\cdot\mid x)$.
Taking $\arg\max$ on both sides yields $\hat{y}(x)=\hat{y}_{\mathrm{cls}}(x)$.
\hfill$\square$

This proposition formalizes that the method cannot degrade predictions on samples where the gate is off.
Therefore, any accuracy change is confined to the gated subset $\{x:g(x)=1\}$.

\subsection{Risk decomposition and sufficient condition for improvement}
\label{sec:appendix:theory:risk}

Let $G$ denote the event $\{g(x)=1\}$ and $\bar{G}$ denote $\{g(x)=0\}$.
By Proposition A.1, the risk difference between the gated method and the classifier decomposes as
\begin{align}
\mathcal{R}(\hat{y})-\mathcal{R}(\hat{y}_{\mathrm{cls}})
&=
\mathbb{E}\!\left[\mathbb{I}[\hat{y}\neq y]-\mathbb{I}[\hat{y}_{\mathrm{cls}}\neq y]\right] \nonumber\\
&=
\mathbb{P}(G)\cdot
\mathbb{E}\!\left[\mathbb{I}[\hat{y}\neq y]-\mathbb{I}[\hat{y}_{\mathrm{cls}}\neq y]\mid G\right].
\label{eq:app:risk_decomp}
\end{align}
Eq.~\eqref{eq:app:risk_decomp} shows that the overall effect is controlled by the conditional effect on the gated subset.

\paragraph{Proposition A.2 (Sufficient condition for risk reduction on $G$).}
Assume that on $G$ the final decision coincides with the similarity decision, i.e., $\hat{y}(x)=\hat{y}_{\mathrm{sim}}(x)$ for all $x\in G$.
If
\begin{equation}
\mathbb{P}(\hat{y}_{\mathrm{sim}}=y\mid G)\;>\;\mathbb{P}(\hat{y}_{\mathrm{cls}}=y\mid G),
\label{eq:app:cond_better}
\end{equation}
then $\mathcal{R}(\hat{y}) < \mathcal{R}(\hat{y}_{\mathrm{cls}})$.
Moreover, if the conditional accuracy gap is at least $\varepsilon>0$, then
\[
\mathcal{R}(\hat{y}) \le \mathcal{R}(\hat{y}_{\mathrm{cls}}) - \mathbb{P}(G)\,\varepsilon.
\]

\paragraph{Proof.}
Under $\hat{y}=\hat{y}_{\mathrm{sim}}$ on $G$, the conditional expectation in Eq.~\eqref{eq:app:risk_decomp} becomes
$\mathbb{P}(\hat{y}_{\mathrm{sim}}\neq y\mid G)-\mathbb{P}(\hat{y}_{\mathrm{cls}}\neq y\mid G)$,
which equals
$-\big(\mathbb{P}(\hat{y}_{\mathrm{sim}}=y\mid G)-\mathbb{P}(\hat{y}_{\mathrm{cls}}=y\mid G)\big)$.
If Eq.~\eqref{eq:app:cond_better} holds, this term is negative, yielding strict risk reduction.
The quantitative bound follows by substituting the $\varepsilon$-gap into Eq.~\eqref{eq:app:risk_decomp}.
\hfill$\square$

Proposition A.2 clarifies what the gate is trying to achieve: it isolates a hard subset where the classifier is less reliable, and on that subset it allows a more reliable similarity-based expert to dominate the decision.

\subsection{When does the fused posterior follow the similarity prediction}
\label{sec:appendix:theory:follow_sim}

Proposition A.2 assumes that the gated decision follows $\hat{y}_{\mathrm{sim}}$.
We now provide a sufficient condition that guarantees this behavior for the linear fusion
$p_{\mathrm{fuse}}=\alpha_{\mathrm{low}}p_{\mathrm{cls}}+(1-\alpha_{\mathrm{low}})p_{\mathrm{sim}}$.

Let $c=\hat{y}_{\mathrm{cls}}(x)$ and $s=\hat{y}_{\mathrm{sim}}(x)$ with $s\neq c$.
Define the pairwise margins
\[
\Delta_{\mathrm{sim}}^{c\leftarrow s}(x)=p_{\mathrm{sim}}(s\mid x)-p_{\mathrm{sim}}(c\mid x),\qquad
\Delta_{\mathrm{cls}}^{c\rightarrow s}(x)=p_{\mathrm{cls}}(c\mid x)-p_{\mathrm{cls}}(s\mid x).
\]
Then
\begin{align}
p_{\mathrm{fuse}}(s\mid x) > p_{\mathrm{fuse}}(c\mid x)
&\Longleftrightarrow
(1-\alpha_{\mathrm{low}})\,\Delta_{\mathrm{sim}}^{c\leftarrow s}(x) \;>\; \alpha_{\mathrm{low}}\,\Delta_{\mathrm{cls}}^{c\rightarrow s}(x).
\label{eq:app:flip_condition}
\end{align}

Our gate enforces (i) low classifier confidence $\gamma_{\mathrm{cls}}(x)<\theta$ and (ii) a decisive similarity posterior via $\Delta_{\mathrm{sim}}(x)>m_{\mathrm{sim}}$.
These imply
\[
\Delta_{\mathrm{cls}}^{c\rightarrow s}(x)\le \gamma_{\mathrm{cls}}(x)<\theta,\qquad
\Delta_{\mathrm{sim}}^{c\leftarrow s}(x)\ge \Delta_{\mathrm{sim}}(x)>m_{\mathrm{sim}}.
\]
Substituting into Eq.~\eqref{eq:app:flip_condition} yields a simple sufficient condition:
\begin{equation}
(1-\alpha_{\mathrm{low}})\,m_{\mathrm{sim}} \;>\; \alpha_{\mathrm{low}}\,\theta
\quad\Longrightarrow\quad
\arg\max_c p_{\mathrm{fuse}}(c\mid x)=s.
\label{eq:app:sufficient_alpha}
\end{equation}
Equivalently,
\begin{equation}
\alpha_{\mathrm{low}} \;<\; \frac{m_{\mathrm{sim}}}{m_{\mathrm{sim}}+\theta}.
\label{eq:app:alpha_bound}
\end{equation}
Eq.~\eqref{eq:app:alpha_bound} links the fusion hyperparameter $\alpha_{\mathrm{low}}$ to the gate threshold $\theta$ and the similarity margin threshold $m_{\mathrm{sim}}$, providing a constructive guideline for selecting $\alpha_{\mathrm{low}}$.

\subsection{A mixture-of-experts view and Bayes-optimal gating}
\label{sec:appendix:theory:moe}

We now formalize the dual-path framework as a two-expert decision system.
Let expert $e\in\{0,1\}$ correspond to $\mathrm{cls}$ (parametric) and $\mathrm{sim}$ (prototype retrieval), respectively.
Each expert induces a deterministic prediction $\hat{y}_e(x)\in\{1,\dots,C\}$.
A gating function selects an expert per input:
\begin{equation}
\hat{y}_g(x)=\hat{y}_{g(x)}(x),\qquad g(x)\in\{0,1\}.
\label{eq:app:hard_gate_def}
\end{equation}
Define the conditional $0$-$1$ risk of expert $e$ at $x$ as
\begin{equation}
r_e(x)=\mathbb{P}\big(\hat{y}_e(x)\neq y\mid x\big)=1-\mathbb{P}\big(\hat{y}_e(x)=y\mid x\big).
\label{eq:app:cond_risk}
\end{equation}

\paragraph{Proposition A.3 (Bayes-optimal gating).}
Among all measurable gating rules $g(\cdot)$, the Bayes-optimal selector that minimizes $\mathcal{R}(\hat{y}_g)$ chooses the expert with smaller conditional risk at each $x$:
\begin{equation}
g^{\star}(x)\in\arg\min_{e\in\{0,1\}} r_e(x).
\label{eq:app:bayes_gate}
\end{equation}
If $r_0(x)\neq r_1(x)$, the optimal gate is unique at $x$.

\paragraph{Proof.}
The expected risk of a gated decision rule is
\[
\mathcal{R}(\hat{y}_g)=\mathbb{E}\big[\mathbb{I}[\hat{y}_{g(x)}(x)\neq y]\big]
=\mathbb{E}\big[\,r_{g(x)}(x)\,\big].
\]
For each fixed $x$, minimizing $r_{g(x)}(x)$ over $g(x)\in\{0,1\}$ yields Eq.~\eqref{eq:app:bayes_gate}.
Since this minimization is pointwise in $x$, the resulting $g^{\star}$ minimizes the expectation.
\hfill$\square$

Proposition A.3 is useful because it characterizes the \emph{ideal} behavior: use retrieval when it is more likely to be correct than the classifier, and otherwise keep the classifier.
However, the conditional risks $r_e(x)$ are not directly observable at test time.
The next subsection explains how confidence, margin, and disagreement form a practical surrogate for this optimal decision.

\subsection{Practical surrogate gating and a regret bound}
\label{sec:appendix:theory:surrogate}

Our gate uses observable statistics from the two posteriors.
Let the expert ``correctness'' probabilities be
\[
a_{\mathrm{cls}}(x)=\mathbb{P}(\hat{y}_{\mathrm{cls}}(x)=y\mid x),\qquad
a_{\mathrm{sim}}(x)=\mathbb{P}(\hat{y}_{\mathrm{sim}}(x)=y\mid x).
\]
Then $r_e(x)=1-a_e(x)$ and the Bayes gate in Eq.~\eqref{eq:app:bayes_gate} selects $\mathrm{sim}$ if $a_{\mathrm{sim}}(x)>a_{\mathrm{cls}}(x)$.

Our implementation approximates $a_e(x)$ using confidence-like surrogates.
Denote $\gamma_{\mathrm{cls}}(x)=\max_c p_{\mathrm{cls}}(c\mid x)$ and $\gamma_{\mathrm{sim}}(x)=\max_c p_{\mathrm{sim}}(c\mid x)$.
The similarity margin $\Delta_{\mathrm{sim}}(x)$ additionally filters out ambiguous retrieval cases.
The disagreement term $D_{\mathrm{JS}}(x)$ prevents intervention when the two posteriors are already similar, where selecting either expert yields limited benefit.

To make this connection explicit, we introduce a mild local calibration assumption.

\paragraph{Assumption A.1 (Local confidence accuracy approximation).}
There exist nonnegative functions $\varepsilon_{\mathrm{cls}}(x)$ and $\varepsilon_{\mathrm{sim}}(x)$ such that
\begin{equation}
\big|a_{\mathrm{cls}}(x)-\gamma_{\mathrm{cls}}(x)\big|\le \varepsilon_{\mathrm{cls}}(x),\qquad
\big|a_{\mathrm{sim}}(x)-\gamma_{\mathrm{sim}}(x)\big|\le \varepsilon_{\mathrm{sim}}(x).
\label{eq:app:local_calib}
\end{equation}
This assumption does not require global perfect calibration.
It only states that, locally, confidence is a usable proxy for correctness up to bounded error.

\paragraph{Proposition A.4 (A sufficient condition for Bayes-consistent expert selection).}
If at a point $x$ the confidence gap satisfies
\begin{equation}
\gamma_{\mathrm{sim}}(x)-\gamma_{\mathrm{cls}}(x)\;>\;\varepsilon_{\mathrm{cls}}(x)+\varepsilon_{\mathrm{sim}}(x),
\label{eq:app:gap_condition}
\end{equation}
then $a_{\mathrm{sim}}(x)>a_{\mathrm{cls}}(x)$ and the Bayes-optimal gate selects the similarity expert at $x$.

\paragraph{Proof.}
From Eq.~\eqref{eq:app:local_calib},
\[
a_{\mathrm{sim}}(x)\ge \gamma_{\mathrm{sim}}(x)-\varepsilon_{\mathrm{sim}}(x),\qquad
a_{\mathrm{cls}}(x)\le \gamma_{\mathrm{cls}}(x)+\varepsilon_{\mathrm{cls}}(x).
\]
If Eq.~\eqref{eq:app:gap_condition} holds, then $a_{\mathrm{sim}}(x) > a_{\mathrm{cls}}(x)$.
\hfill$\square$

Proposition A.4 explains the role of our thresholds: enforcing $\gamma_{\mathrm{cls}}(x)<\theta$ and $\gamma_{\mathrm{sim}}(x)>\beta$ implicitly encourages a large confidence gap, while the margin constraint $\Delta_{\mathrm{sim}}(x)>m_{\mathrm{sim}}$ makes $\gamma_{\mathrm{sim}}$ more reliable by excluding uncertain retrieval cases.
Moreover, the divergence constraint $D_{\mathrm{JS}}(x)>\delta$ focuses corrections on inputs where the two branches disagree meaningfully; when posteriors are similar, selection has limited effect and leaving the classifier unchanged is a safe default.

Finally, we provide a regret-style bound relative to the Bayes gate, expressed via the probability of mis-ranking the two experts.

\paragraph{Proposition A.5 (Regret bound via expert mis-ranking).}
Let $g^{\star}$ be the Bayes-optimal gate in Eq.~\eqref{eq:app:bayes_gate}$.$
For any gate $\tilde{g}$, define the mis-ranking event
\[
\mathcal{E}=\{x:\ r_{\tilde{g}(x)}(x) > r_{g^{\star}(x)}(x)\}.
\]
Then the excess risk satisfies
\begin{equation}
\mathcal{R}(\hat{y}_{\tilde{g}})-\mathcal{R}(\hat{y}_{g^{\star}})
\;\le\;
\mathbb{P}(\mathcal{E}).
\label{eq:app:regret_bound}
\end{equation}

\paragraph{Proof.}
For each $x$, $r_{\tilde{g}(x)}(x)-r_{g^{\star}(x)}(x)\le 1$ and is strictly positive only on $\mathcal{E}$.
Thus
\[
\mathcal{R}(\hat{y}_{\tilde{g}})-\mathcal{R}(\hat{y}_{g^{\star}})
=\mathbb{E}\big[r_{\tilde{g}(x)}(x)-r_{g^{\star}(x)}(x)\big]
\le \mathbb{E}\big[\mathbb{I}[x\in\mathcal{E}]\big]=\mathbb{P}(\mathcal{E}).
\]
\hfill$\square$

Eq.~\eqref{eq:app:regret_bound} indicates that the effectiveness of a practical gate depends on how often it selects the worse expert.
Our training choices (supervised contrastive geometry, teacher stabilization, and multi-prototype modeling) are designed to reduce this mis-ranking probability by making the similarity posterior reliable precisely on the hard subset where the classifier confidence is low.

\subsection{Connection back to the proposed gated fusion}
\label{sec:appendix:theory:connection}

The theoretical MoE analysis above is stated for hard selection (Eq.~\eqref{eq:app:hard_gate_def}), while our method applies a conservative fusion (Eq.~\eqref{eq:app:final_posterior_def}).
When Eq.~\eqref{eq:app:alpha_bound} holds, the fused decision on gated samples matches the similarity prediction, and Propositions A.2--A.5 apply directly.
When Eq.~\eqref{eq:app:alpha_bound} does not strictly hold, convex fusion still reduces the chance of abrupt decision flips caused by imperfect retrieval evidence, which is desirable under residual calibration error.
In both cases, the gate localizes the intervention to ambiguous inputs and uses reliability constraints to approximate the Bayes-optimal expert selection behavior.

\subsection{Algorithm}

\begin{algorithm}[t]
\caption{Teacher-guided Dual-path Inference Framework}
\label{alg:method}
\begin{algorithmic}[1]
\State \textbf{Input:} Test image $x$, classifier parameters $\theta$, $\phi$, teacher parameters $\bar{\theta}$, $\bar{\phi}$, prototype bank $\mathcal{P}$.
\State Compute teacher embedding $\bar{z}(x) \gets \frac{g_{\bar{\phi}}(f_{\bar{\theta}}(x))}{\|g_{\bar{\phi}}(f_{\bar{\theta}}(x))\|_2}$.
\For{each class $c$}
    \State Compute similarity scores $s_{c,k}(x) \gets \bar{z}(x)^{\top} p_{c,k}$.
\EndFor
\State Compute class score $q_c(x) \gets \log \sum_{k=1}^{K} \exp\left(\kappa s_{c,k}(x)\right)$.
\State Compute similarity posterior:
\[
p_{\mathrm{sim}}(y=c \mid x) = \frac{\exp(q_c(x)/\tau_{\mathrm{sim}})}{\sum_{j=1}^{C} \exp(q_j(x)/\tau_{\mathrm{sim}})}.
\]
\State \textbf{Confidence-gated Fusion:}
\State Compute classifier confidence $\gamma_{\mathrm{cls}}(x)$, entropy $H_{\mathrm{cls}}(x)$, and prototype reliability $\gamma_{\mathrm{sim}}(x)$ according to Eq.~\ref{eq:sim_posterior}-\ref{eq:entropy_cls}.
\State Compute similarity margin $\Delta_{\mathrm{sim}}(x)$ and divergence $D_{\mathrm{JS}}(x)$.
\State Compute gate $g(x)$ based on thresholds:
    \[
    g(x)
    =
    \mathbb{I}\!\left[\gamma_{\mathrm{cls}}(x)<\theta\right]\cdot
    \mathbb{I}\!\left[\gamma_{\mathrm{sim}}(x)>\beta\right]\cdot
    \mathbb{I}\!\left[\Delta_{\mathrm{sim}}(x)>m_{\mathrm{sim}}\right]\cdot
    \mathbb{I}\!\left[D_{\mathrm{JS}}(x)>\delta\right]\cdot
    \mathbb{I}\!\left[\hat{y}_{\mathrm{cls}}\neq \hat{y}_{\mathrm{sim}}\right],
    \label{eq:gate_full}
    \]
\If{$g(x) = 1$}
    \State Apply fusion:
    \[
    p_{\mathrm{fuse}}(\cdot \mid x) = \alpha_{\mathrm{low}} \cdot p_{\mathrm{cls}}(\cdot \mid x) + (1-\alpha_{\mathrm{low}}) \cdot p_{\mathrm{sim}}(\cdot \mid x).
    \]
\Else
    \State Keep classifier prediction unchanged.
\EndIf
\State \textbf{Output:} Final prediction $p(\cdot \mid x)$.
\end{algorithmic}
\end{algorithm}

\end{document}